# SINERGY: A Linear Planner Based on Genetic Programming

Ion Muslea

Information Sciences Institute / University of Southern California
4676 Admiralty Way
Marina del Rey, CA 90292 (USA)
muslea@isi.edu

**Abstract.** In this paper we describe SINERGY, which is a highly parallelizable, linear planning system that is based on the genetic programming paradigm. Rather than reasoning about the world it is planning for, SINERGY uses artificial selection, recombination and fitness measure to generate linear plans that solve conjunctive goals. We ran SINERGY on several domains (e.g., the briefcase problem and a few variants of the robot navigation problem), and the experimental results show that our planner is capable of handling problem instances that are one to two orders of magnitude larger than the ones solved by UCPOP. In order to facilitate the search reduction and to enhance the expressive power of SINERGY, we also propose two major extensions to our planning system: a formalism for using hierarchical planning operators, and a framework for planning in dynamic environments.

## 1 Motivation

Artificial intelligence planning is a notoriously hard problem. There are several papers [Chapman 1987, Joslin and Roach 1989, Bylander 1992] that provide in-depth discussions on the complexity of AI planning, and it is generally accepted that most non-trivial planning problems are at least NP-complete. In order to cope with the combinatorial explosion of the search problem, AI researchers proposed a wide variety of solutions, from search control rules [Weld 1994, Etzioni 1993, Minton 1996] to abstraction and hierarchical planning [Knoblock 1991 and 1994] to skeletal planning [Friedland 1985]. However, even though the above-mentioned techniques may dramatically narrow the search space, there is no guarantee that the corresponding planning algorithms will gracefully scale up for real-world problems.

As a reaction to the shortcomings of the traditional planners, during the last couple of years we witnessed the occurrence of new type of planning systems: the stochastic planners. This new approach to AI planning trades in the completeness of the planner for the speed up of the search process. Planners like SatPlan [Kautz and Selman 1996] or PBR [Ambite and Knoblock 1997] are at least one order of magnitude faster than the classic planning systems, and they are also capable of handling significantly larger problem instances.

In this paper we present SINERGY, which is a general-purpose, stochastic planner based on the genetic programming paradigm [Koza 1992]. Genetic Programming (GP) is an automatic programming technique that was introduced as an extension to the genetic algorithms (GA) [Holland 92], and it uses evolution-like operations (e.g., reproduction and cross-over) to generate and manipulate computer programs. In [Koza 1992 and 1994, Spector 1994, Handley 1994], the authors used GP to solve several problems that are similar to the ones encountered in AI planning (e.g., the Sussman anomaly, the robot navigation problem, and an unusual variant of the block world problem). Even though their domain-specific solutions cannot be considered general-purpose planning systems, the experimental results showed that GP has great potential for solving large instances of traditional AI planning problems.

Based on the encouraging results obtained by both stochastic planners and GP-based problem solving techniques, we decided to formalize and fully-implement a general-purpose AI planner that relies on the genetic programming paradigm. Rather than reasoning about the world it is planning in, SINERGY uses artificial selection, recombination and fitness measures to generate linear plans that solve conjunctive goals. We must emphasize that SINERGY has an expressive power equivalent to the one offered by UCPOP: it provides conditional effects, disjunctive preconditions, and both universal and existential quantifiers. We tested our planner on several domains, and the experimental results show that SINERGY is capable of handling problem instances that are one to two orders of magnitude larger than the ones solved by UCPOP.

## 2  Genetic Programming

Genetic Programming represents a special type of genetic algorithm in which the structures that undergo adaptation are not data structures, but hierarchical computer programs of different shapes and sizes. The GP process starts by creating an initial population of randomly-generated programs and continues by producing new generations of programs based on the Darwinian principle of "the survival of the fittest". The automatically-generated computer programs are expressed as function composition, and the main breeding operations are *reproduction* and *cross-over*. By reproduction we mean that a program from generation *i* is copied unchanged within generation *i+1*, while the cross-over takes two parent-programs from generation *i*, breaks each of them in two components, and adds to generation *i+1* two children-programs that are created by combining components coming from different parents.

In order to create a GP-based application, the user has to specify a *set of building-blocks* based on which the population of programs is constructed, and an *evaluation function* that is used to measure the fitness of each individual program. There are two types of primitive elements that are used to build a program: *terminals* and *functions*. Both terminals and functions can be seen as LISP-functions, the only difference between them consisting of the number of arguments that they are taking: terminals are not allowed to take arguments, while functions take at least one argument. The *individuals* generated by the GP system represent computer *programs* that are built by *function composition* over the set of terminals and functions. Consequently, GP imposes the *closure property*: any value returned by a function or a terminal must represent a valid input for any argument of any function in the function set.

As we have already mentioned, the GP problem specification must include a domain-specific *fitness evaluation function* that is used by the GP system to estimate the "fitness" of each individual of a generation. More specifically, the fitness function takes as input a GP-generated program *P*, and its output represents a measure of how appropriate *P* is to solve the problem at hand (in this paper, lower fitness values mean better programs). In order to estimate the fitness of an GP-generated program, the evaluation function uses a set of *fitness cases.* Each fitness case is a tuple <*in-values, desired-out-values*> that has a straightforward meaning: for the given *in-values*, a "perfectly fit" program should generate the *desired-out-values*. In terms of AI planning, the *in-values* represent the initial world status, while the *desired-out-values* can be seen as the goals to be achieved.

Once the terminals, functions, fitness cases and fitness function are specified, the user has only to select a few running parameters (e.g., number of programs per generation, number of generations to be created, maximum size of a program) and to let the GP system evolve the population of programs. Both cross-over and reproduction are performed on randomly chosen individuals, but they are biased for highly fit programs. Such an approach has two major advantages: on one hand, the "highly fit" bias leads to the potentially fast discovery of a solution, while on the other hand, GP is capable of avoiding local minima by also using in the breeding process individuals that are less fit than the "best" offsprings of their respective generations. Although GP problem solvers are not complete (i.e., if there is a solution to the problem, GP-based systems are not guaranteed to find it), the experimental results show that they are usually able to find close-to-optimal solutions for a wide range of problems, from robotics to pattern recognition to molecular biology.

## 3   Planning as Genetic Programming

Even though there are several exceptions (e.g., NOAH [Sacerdoti 1975]), the vast majority of the AI planners define their planning actions in a declarative manner that is based on the one used by STRIPS [Fikes and Nilsson 1971]. In contrast, SINERGY takes a different approach and requires a procedural description of the planning operators. SINERGY relies on a procedural description of the planning actions because its underlaying, GP-based problem solver has to execute in simulation each plan in order to estimate its fitness. However, as Table 1 shows, the different nature of traditional AI planners and our GP-based planner (i.e., reasoning about plans *vs.* generating and executing genetically-created plans) does not prevent the two categories of planning systems to have similar interfaces.

**Table 1.** Interface Comparison: Traditional AI Planning Systems *vs.* SINERGY

|  | Traditional AI Planners | SINERGY |
|---|---|---|
| Input | - initial state<br>- set of goals<br>- set of operators<br>- additional information (search control rules, memory of plans, etc.) | - initial state<br>- set of goals<br>- set of operators<br>- additional information (fitness evaluation functions) |
| Output | Plan: a (partially ordered) sequence of fully instantiated operations | Plan: a *linear* sequence of fully instantiated operations |

In order to find a solution for the problem at hand, SINERGY uses an approach resembling the one described in [Kautz and Selman 1996]: it converts an AI planning problem $P_1$ to a problem $P_2$ of a different nature, it solves $P_2$ based on a stochastic approach, and it converts the result to a solution for $P_1$. However, while Kautz and Selman turned the AI planning problem to an equivalent satisfiability problem, SINERGY converts the AI planning problem to a GP problem (Figure 1).

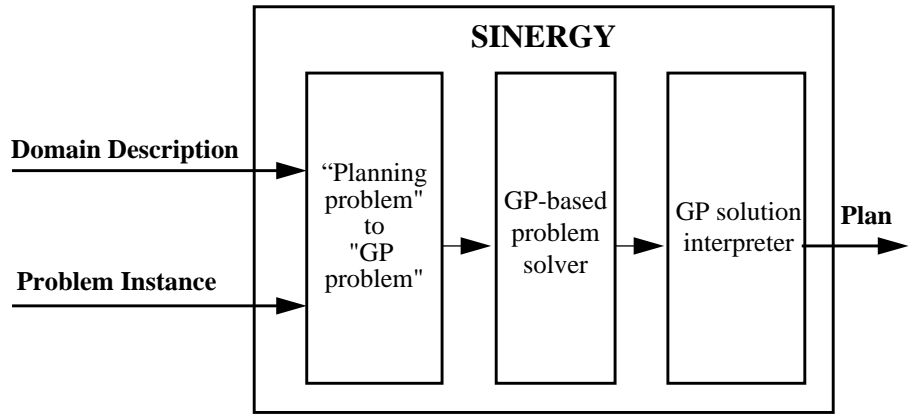

**Fig. 1.** The SINERGY approach to AI planning

The SINERGY approach to solving AI planning problems has three main advantages. First of all, GP problem solving is by its very nature a highly parallelizable process. The most expensive operation in terms of CPU-time is the fitness evaluation, and it is easy to see that a GP-based planner scales up gracefully to any number of parallel processors: the fitness evaluations of different plans represent completely independent processes that can by easily performed on different processors. Second, the SINERGY approach to AI planning facilitates problem solving in dynamic environments. As each plan simulation has to be executed step by step, we can create a framework that would allow the dynamic update of the planning environment after the execution of each operation in the plan (e.g., for the robot navigation problem, we can define mobile obstacles that move on predefined trajectories). Third, SINERGY provides a flexible way to express goal priorities by requiring the definition of a fitness evaluation function for each goal-predicate (see the next section). Domain implementers can use weight-factors to express the relative importance of each goal type, and, consequently, plans that solve a larger number of higher-importance goals will have a better fitness value and will be preferred during the selection and recombination process.

GP-based problem solvers rarely find optimal solutions for the problems at hand, but the above mentioned advantages together with SINERGY's ability to find close-to-optimal solutions for large, complex problem instances makes the GP-based approach to AI planning an alternative to be taken into account and a serious candidate for further research investigations.

## 4  Specifying a Planning Domain for SINERGY

In this section we analyze the domain representation features offered by SINERGY. In order to make this process as intuitive as possible, we will present an example of a domain description for the briefcase problem (BP). In a few words, we can define BP as follows: given a number of briefcases ($B=\{B_1, B_2, ... , B_n\}$), objects ($O=\{O_1, O_2, ...., O_m\}$), and locations ($L=\{L_1, L_2, ..., L_k\}$), we want to deliver each object $O_i$ to its respective destination $L^i_j$. In order to solve the problem, the planning system can use one of the following operators:

a) *(move-briefcase a-briefcase a-location)*: moves *a-briefcase*, together with all the objects in it, from the current location to *a-location*.

b) (put-in *an-object a-briefcase*): if *an-object* and *a-briefcase* are at the same location, puts *an-object* in *a-briefcase.*

c) (take-out *an-object*): if SINERGY finds that *an-object* is in a briefcase, say *B*, it takes *an-object* out of *B*.

In SINERGY, any specification of the planning domain has four main descriptive sections: concepts, predicates, planning operators, and auxiliary functions. Figure 2 shows the whole domain specification for the briefcase problem. *Concept definitions* are self-explanatory, and they simply introduce the *types* of entities that can be manipulated by the planner. We will focus first on *predicate specifications*, which require several important comments, and we will continue our discussion by analyzing the *planning operators* and the *auxiliary functions*.

Any SINERGY predicate can be used both to describe the state of the world (e.g., a given object is *at* a given location or *in* a briefcase) and to specify a goal (e.g., the object *obj* must be transported *at* a given location or put *in* a briefcase). As the planning problem is converted to a GP one, each predicate must have a corresponding fitness evaluation function. Even though in our example the fitness evaluation functions are extremely simple (i.e., they just count the number of unsatisfied goals), a fitness evaluation function might be arbitrarily complex: it can use heuristics like the Manhattan distance (e.g., for the robot navigation problem), or it might include weight-factors that would allow the user to define a hierarchy of the goals based on their relative importance (i.e., the more important a type of a goal, the larger its weight-factor will be).

As we have already seen, SINERGY does not reason about plans, but instead uses artificial selection and recombination to create plans that are likely to have a better fitness measure than the ones in the previous generation. In order to evaluate the fitness of a given plan *P*, which is a linear sequence of fully instantiated planning operators $p_1, p_2, ..., p_l$, SINERGY sets the world status to the given initial state $S_0$, and it successively simulates the execution of each operator $p_i$ in the plan. At the end of the plan execution, the planner computes the fitness of *P* based on the formula

$$\sum_{i=1}^{n} FitnessEvaluationFunction_{Predicate_i}(CurrentState, Goals)$$

```lisp
;; Concept definitions.
(defvar *concepts* '(object briefcase location))

;;Predicate definitions: pairs <predicate fitness-eval-fct>.
(defvar *predicates* '((in in-fitness) (at at-fitness)))

(defun in-fitness( relevant-goals )
 (number-of-unsatisfied-goals relevant-goals))

(defun at-fitness( relevant-goals )
 (number-of-unsatisfied-goals relevant-goals))

;; Operator definitions (names and number of arguments).
(defvar *planning-operators*
       '( (move-briefcase 2) (take-out 1) (put-in 2) ))

(defun put-in( arg-1 arg-2 )
 (let ((an-object (convert-to arg-1 'object))
       (a-briefcase (convert-to arg-2 'briefcase)))
  (when ;; PRECONDITIONS
     (equal (get-location an-object) (get-location a-briefcase))
     ;; EFFECTS
     (add-fact `(in ,an-object ,a-briefcase))))
 arg-1)

(defun take-out( arg-1 )
 (let ((an-object (convert-to arg-1 'object)))
   (when ;; PRECONDITIONS
     (is-fact `(in ,an-object ,(get-briefcase an-object)))
     ;; EFFECTS
     (delete-fact `(in ,an-object ,(get-briefcase an-object)))))
 arg-1)

(defun move-briefcase ( arg-1 arg-2 )
 (let((a-briefcase (convert-to arg-1 'briefcase))
      (a-location (convert-to arg-2 'location)))
  (when ;; PRECONDITIONS
    (not (is-fact `(at ,a-briefcase ,a-location)))
    ;; EFFECTS
    (for-all 'object #'do-move (list a-briefcase a-location))

    (delete-fact `(at ,a-briefcase ,(get-location a-briefcase)))
    (add-fact `(at ,a-briefcase ,a-location))))
 arg-1)

;; Auxiliary functions.
(defun do-move ( obj br to )
  (when (is-fact `(in ,obj ,br))
        (delete-fact `(at ,obj ,(get-location obj)))
        (add-fact `(at ,obj ,to))))

(defun get-location (object)
  (find-attribute-value '?location `(at ,object ?location)))

(defun get-briefcase (object)
  (find-attribute-value '?briefcase `(in ,object ?briefcase)))
```

**Fig. 2.** The Domain Description for the Briefcase Problem.

*Planning operators* are defined as a collection of Lisp functions that are used as terminals and functions by the GP problem solver. In order to keep track of the world status during a plan execution, SINERGY provides the predefined functions *add-fact* and *delete-fact* that must be used by the planning operators whenever they change the world status. As a direct consequence of the procedural definition of the planning operators, the use of disjunctive preconditions and conditional effects is a trivial task in SINERGY. Furthermore, our planning system offers additional features that make its domain description language extremely powerful and expressive. First, SINERGY allows users to define *auxiliary functions* that can be invoked within the operators. For instance, in our BP example, *get-location* and *get-briefcase* are used to determine the current location, respectively the briefcase that contains a given object. Second, SINERGY provides both universal and existential quantifiers that have the following syntax:

> **for-all** *concept-name auxiliary-function-name additional-arguments*
>
> **exists** *concept-name auxiliary-function-name additional-arguments*

In the definition of the *move-briefcase* operator we used the universal quantifier to apply an action to all instances of the *object* concept, but both types of quantifiers can also be used in the goal specification of any problem instance.

SINERGY also provides a domain-independent solution to a major problem related to the different nature of AI planning and GP: planning operators are "strongly typed" (i.e., each argument of an operator must be of a well-defined, pre-established type), while the GP functions introduced in [Koza 1992] are "typeless" because they always rely on the *closure property*, which ensures that any value returned by a function or terminal represents a valid actual parameter for any function in the function set. In order to solve this problem, we used the following approach: in addition to the user-defined terminals (i.e., planning operators that take no arguments), the GP problem solver uses a supplementary set of terminals $T_s = \{ t_1, t_2, ...t_n \}$. For a given problem instance, SINERGY automatically generates $T_s$ in such a way that the value returned by a terminal $t_i$ can be converted to a unique object name of *any* type defined within the planning universe. Consequently, each planning operator must convert its parameters to objects of the desired type by invoking the SINERGY-provided *convert-to* function.

Finally, for a given planning domain, the user has to specify the instance of the problem to be solved (Figure 3). The variable *\*concept-instances\** defines the domain objects of each type (e.g., briefcases, locations, or objects-to-be-moved), while *\*init-state\** and *\*goal-state\** are used by SINERGY to generate a fitness case for the GP problem solver. Based on the information in *\*concept-instances\**, SINERGY also creates the internal data structures that allow the *convert-to* function to uniformly map GP entities to valid domain objects.

## 5 An Example of Plan Evolution Based on Genetic Recombination

In order to better understand how SINERGY creates new plans from the existing ones, we will analyze an example of plan construction that is based on the GP crossover operation. Let us suppose that SINERGY tries to solve the BP instance that is presented in Figure 3. Based on the BP domain description, our planner generates an empty set of GP terminals and three GP functions: *take-out*, *put-in*, and *move-*

*briefcase* (*take-out* has one argument, while *put-in* and *move-briefcase* require two arguments).

```
(defvar *concept-instances*
 '( (object (o1 o2)) (briefcase (b1)) (location (l1 l2 l3))))
(defvar *init-state* '((at o1 l1)(at o2 l3) (at b1 l1)))
(defvar *goal-state* '((at o1 l2))))
```

**Fig. 3.** The Definition of a Simple BP Instance.

After analyzing the domain description, SINERGY examines the problem instance and generates the additional set of terminals $T_s$ that was briefly discussed in the previous section. As it is beyond the scope of this paper to explain the algorithms based on which $T_s$ and the corresponding data structures are generated (for details see [Muslea 1997]), let us accept without further proof that for the given BP instance SINERGY creates a set of six additional terminals $T_s = \{t_1, t_2, t_3, t_4, t_5, t_6\}$, and the function *convert-to* provides the mappings described in Table 2. The information from Table 2 must be interpreted as follows: the function call *(convert-to ($t_2$) 'location)* returns the location name $l_2$, while the function call *(convert-to ($t_6$) 'object)* returns the object name $o_2$. Note that for concepts that have a unique instance (e.g., *briefcase*) all terminals are mapped to the unique object name, while for concepts with several instances, each object name corresponds to the *same* number of distinct terminals from $T_s$.

**Table 2.** CONVERT-TO mapping of terminals to domain objects.

| ***CONVERT-TO*** | location | object | briefcase |
|---|---|---|---|
| $t_1$ | $l_1$ | $o_1$ | $b_1$ |
| $t_2$ | $l_2$ | $o_2$ | $b_1$ |
| $t_3$ | $l_3$ | $o_1$ | $b_1$ |
| $t_4$ | $l_1$ | $o_2$ | $b_1$ |
| $t_5$ | $l_2$ | $o_1$ | $b_1$ |
| $t_6$ | $l_3$ | $o_2$ | $b_1$ |

Now let us suppose that in Generation 1 the GP system creates the two random programs *P1* and *P2* presented in Table 3. Both *P1* and *P2* are expressed as function compositions over the sets of GP terminals and functions. During the plan-simulation phase, the GP problem solver executes the programs *P1* and *P2* by starting with the inner-most functions (i.e., the terminals $t_1$ and $t_2$) and ending with the outmost ones (e.g., *take-out*, respectively *move-briefcase*). If we use the *convert-to* function to map each occurrence of the terminals $t_1$ and $t_2$ to *object names* of the types specified in each planning operator, the programs *P1* and *P2* can be interpreted as the equivalent linear plans *<(put-in $o_1$ $b_1$), (take-out $o_1$)>*, respectively *<(take-out $o_2$), (move-briefcase $b_1$ $l_2$)>*. As none of these two plans satisfies the goal *(at $o_1$ $l_2$)*, the GP problem solver will create a new generation of programs.

In Table 3, we assumed that in order to create the second generation, the GP system applies the cross-over operator to the plans *P1* and *P2*. As the recombination process arbitrarily breaks each parent in two components, let us suppose that in the current example the GP system chooses to interchange the high-lighted portions of *P1* and *P2*, which leads to the creation of the new plans *C1* and *C2*. Even though the child-plan *C1* is useless and contains redundant operators, the plan *C2* represents a solution to our BP instance because after its simulated execution the goal *(at $o_2$ $l_2$)* is satisfied. As a final note, we must emphasize that most of the GP-generated plans are similar to *C1* in the sense that they include redundant operators, and, in many cases, the preconditions of the fully-instantiated operators are not satisfied (e.g., during the simulation of *C1*, none of the operators can be actually executed because the object $o_1$ is not in the briefcase $b_1$). However, after multiple recombinations, plan fragments might fit together in such a way that a newly created plan solves the problem at hand.

**Table 3.** Programs *P1* and *P2* are recombined into *C1* and *C2*.

|    | GP-generated Programs | GP Functions Executed During Plan Simulation | Equivalent Linear AI Plans |
|----|-----------------------|----------------------------------------------|----------------------------|
| **P1** | (take-out **(put-in ($t_1$) ($t_2$)))** | 1: (put-in ($t_1$) ($t_2$))<br>2: (take-out ($t_1$)) | 1: (put-in $o_1$ $b_1$)<br>2: (take-out $o_1$) |
| **P2** | (move-briefcase **(take-out ($t_2$))** ($t_2$)) | 1: (take-out ($t_2$))<br>2: (move-briefcase ($t_2$) ($t_2$)) | 1: (take-out $o_2$)<br>2: (move-briefcase $b_1$ $l_2$) |
| **C1** | (take-out **(take-out ($t_2$)))** | 1: (take-out ($t_2$))<br>2: (take-out ($t_2$)) | 1: (take-out $o_2$)<br>2: (take-out $o_2$) |
| **C2** | (move-briefcase **(put-in ($t_1$) ($t_2$))** ($t_2$)) | 1: (put-in ($t_1$) ($t_2$))<br>2: (move-briefcase ($t_1$) ($t_2$)) | 1: (put-in $o_1$ $b_1$)<br>2: (move-briefcase $b_1$ $l_2$) |

## 6 Experimental Results

In order to have an accurate image of SINERGY's capabilities, we ran our planner on three different domains: the single robot navigation problem (RNP), the 2-robot navigation problem (2RNP), and the briefcase problem (BP). In this paper, we define RNP as follows: given a rectangular *m*-by-*n* table T with *k* blocks located in its grid-cells, a robot R must navigate from its current position CP to the desired position DP. In order to reach DP, the robot can use any of the eight available operations: *move* (north/south/east/west) to an unoccupied neighboring cell, or *push* (north/south/east/west) a block located in a neighboring cell X to an empty cell Y that is right behind X. 2RNP is similar to RNP, but it requires that both robots reach their respective destination. RNP is an extremely hard problem because of the high level of interaction among the operators' effects (i.e., if the robot pushes a block to an inappropriate location, it may bring the universe into a status from where the problem is not solvable anymore), and 2RNP is even harder because the two robots might have conflicting goals.

For all our experiments, we ran SINERGY on a maximum of 1000 generations of 200 individuals each. The population size is extremely small in terms of GP problem solvers, but due to our hardware limitations (we ran the experiments on a single-processor, non-dedicated SUN-4 machine) we could not afford to consistently use a larger population. However, we made a few experiments on 50 generations of 2000 individuals, and the solution was found in significantly fewer generations because the larger initial population increases the chances of finding well-fit plans from the very first generation. For each of the three domains mentioned above, we ran SINERGY on more than 100 problem instances, and our results for the most difficult instances are presented in Tables 4, 5 and 6.

**Table 4.** Results for the Single Robot Navigation Problem

| Instance | Description | Generation of First Solution |
| --- | --- | --- |
| RNP-1 | 4x4 table, 6 obstacles | 1 |
| RNP-2 | another 4x4 table, 6 obstacles | 1 |
| RNP-3 | 8x8 table, 18 obstacles (requires at least 6 PUSH-es) | 10 |
| RNP-4 | 50x50 table, no obstacles | 150 |
| RNP-5 | 100x100 table, no obstacles | 317 |
| RNP-6 | 100x100 table, same obstacles positions as in RNP-1 | 354 |
| RNP-7 | 100x100 table, same obstacles positions as in RNP-2 | 333 |
| RNP-8 | 100x100 table, same obstacles positions as in RNP-3 | 353 |

In Table 4, we show the results of running SINERGY on several difficult instances of RNP that UCPOP was not able to solve. Based on the experimental results, we can make several important observations. First, SINERGY is capable of solving hard problem instances that are two orders of magnitude larger than the ones solved by UCPOP (note: for both RNP and 2RNP we used a fitness evaluation function based of the Manhattan distance between the position reached by the robot and the desired position). For instance, in RNP-8 we have a 100-by-100 table with 18 blocks that completely obstruct the way from the initial position (0,0) to the destination (99,99), and the robot must perform two complicated sequences of PUSH operations (one to get away from its initial location, and another one to make its way toward the final destination). Second, SINERGY is capable of creating better plans (i.e., closer to the optimal solution) once it finds a first solution. For example, SINERGY found a first solution for RNP-6 at generation 354 (it had 293 operations, while the optimal plan required only 206 actions), but by the time it reached generation 867, our planner kept improving the plan and was able to deliver a solution with only 216 operations. Finally, SINERGY is especially well fit to solve hard problem instances. While on easy instances, like small tables with no blocks, UCPOP is faster than SINERGY, the GP-based approach is more appropriate for problem instances that have a higher level of difficulty.

**Table 5.** Results for the 2-Robot Navigation Problem

| Instance | Description | Generation of First Solution |
|---|---|---|
| 2RNP-1 | 8x8 table, no obstacles | 9 |
| 2RNP-2 | 8x8 table, 2 obstacles | 14 |
| 2RNP-3 | 8x8 table, 5 obstacles | 24 |
| 2RNP-4 | 8x8 table, 10 obstacles | 49 |
| 2RNP-5 | 100x100 table, 18 obstacles | - |

The experiments presented in Table 5 are of a different nature: for 2RNP, we tested SINERGY's performance on domain specifications that are totally unfit for the GP approach. That is, rather than extending the domain representation for RNP such that both the *move* and *push* operators take an additional parameter that denotes the robot-to-perform-the-action, we decided to use the same set of operators, and to interpret a generated plan as follows: by default, all the even operators are performed by $robot_1$, while the odd ones are executed by $robot_2$. It is easy to see that such a representation is not fit for GP-based problem solvers: if the recombination of two parent-plans $P_1 = <a_1, a_2, ..., a_n>$ and $P_2 = <b_1, b_2, ..., b_m>$ generates the children-plans $C_1 = <a_1, a_2, ..., a_{2k+1}, b_{2l+1}, b_{2l+2}, ..., b_m>$ and $C_2 = <b_1, b_2, ..., b_{2l}, a_{2k+2}, a_{2k+3}, ..., a_n>$, the operators $b_{2l+1}, b_{2l+2}, ..., b_m$ from $C_1$ and $a_{2k+2}, a_{2k+3}, ..., a_n$ from $C_2$ will be executed by different robots than the ones that executed them within the original plans $P_1$ and $P_2$. However, despite the inappropriate encoding of the plans, SINERGY is capable of finding a solution for all medium-size test-instances.

**Table 6.** Results for the Briefcase Problem

| Instance | Description | Generation of First Solution |
|---|---|---|
| BP-1 | 4 objects, 5 locations, 1 briefcase | 59 |
| BP-2 | 5 objects, 5 locations, 1 briefcase | 42 |
| BP-3 | 5 objects, 5 locations, 5 briefcases | 42 |
| BP-4 | 10 objects, 10 locations, 1 briefcase | 66 |
| BP-5 | 10 objects, 10 locations, 2 briefcases | 68 |
| BP-6 | 10 objects, 10 locations, 5 briefcases | 136 |
| BP-7 | 10 objects, 10 locations, 10 briefcases | - |

Finally, our last set of tests was performed for the briefcase problem (Table 6). We used the domain specification presented in Figure 2, and the experimental results are similar to the ones obtained for the RNP domain: on easy instances (e.g., all objects must be transported to the same place and are initially stored at the same location) UCPOP solves the problem faster than SINERGY, but on complex problem instances (e.g., large number of objects, each of them being initially located at a different location) SINERGY is still capable of solving the problem, while UCPOP is unable to cope with the increased level of difficulty. By analyzing the results of SINERGY in the three domains that we considered, we can conclude that our planner significantly outperforms UCPOP for all difficult test-instances, but it is slower than UCPOP on most of the easy ones.

## 7   Future Work

We plan to extend SINERGY by adding two major features. First, we would like to facilitate the search reduction by introducing hierarchical planning operators. The new version of SINERGY would allow users to create several levels of abstraction, each of them being described by a distinct set of operators and predicates. For instance, if we consider a combination of the briefcase problem and the robot navigation problem (*BRNP*), at the higher level of abstraction each robot could perform the three BP operations (*put-in, take-out,* and *move-to*), and SINERGY would not be concerned with any navigation details. Once the planner finds a solution at the higher level of abstraction, it translates each fully-instantiated *(move-to robot$_i$ location$_j$)* operator into a goal *(at robot$_i$ x$_j$ y$_j$)* that must be solved at a lower level of abstraction. In order to satisfy the newly generated goals, SINERGY must solve the navigation problem based on a different set of operators (e.g. *move*, *rotate*, or *push*). The use of hierarchical planning operators might be extremely beneficial for domains like *BRNP*, in which achieving the goal *(at object 100 13)* involves a long sequence of *move* and *push* operators, followed by a single *take-out* action.

Second, we would like to allow users to define planning problems that involve dynamic environments. For instance, in the robot motion problem, we could define two distinct types of obstacles: fixed blocks and mobile blocks. Fixed blocks could change their positions only if they are pushed by a robot, while mobile blocks would be continuously changing their positions based on predefined trajectories. As candidate plans are executed in simulation, after performing each planning operation SINERGY could update the position of the mobile blocks based on the predefined trajectory functions. We believe that the ability of SINERGY to plan in dynamic environments would represents a major advantage of our approach because most real-world problems must be solved in dynamic environments, and traditional planners are generally unable to cope with such environments.

## 8   Conclusions

The major contribution of this paper consists of providing a domain-independent mapping of any AI planning problem into an equivalent GP problem. In final analysis, we can conclude that SINERGY is a general-purpose AI planning system that is capable of solving large, complex problem instances. By supporting disjunctive preconditions, conditional effects, and both existential and universal quantifiers, SINERGY provides users with a domain description language that has an expressive power equivalent to the one offered by UCPOP. Our initial results show that SINERGY outperforms UCPOP on all the difficult examples it was tested on. Furthermore, the highly parallelizable nature of GP makes us believe that running SINERGY on a relatively low-power, parallel machine would allow our planner to easily solve problem instances at least two orders of magnitude larger than the ones presented in this paper.

Even though SINERGY is an incomplete planner that does not generally find the optimal solution for a given problem, its practical ability to solve complex problems and to improve the quality of its initial solution makes it a valuable tool for dealing with hard problems. We plan to enhance SINERGY by adding hierarchical operators and a formalism for handling dynamic universes, and we expect the new version to provide both a faster planning process and significantly more expressive power.